# EyeGPT: Ophthalmic Assistant with Large Language Models

Xiaolan Chen, MD[1#], Ziwei Zhao, MD[1#], Weiyi Zhang, MS[1#], Pusheng Xu, MD[4#], Le Gao, MD[4], Mingpu Xu, MD[1], Yue Wu, MD[5,6], Yinwen Li, MD[5,6], Danli Shi, MD, PhD[1,2,3*], Mingguang He, MD, PhD[1,2,3,4]

[1] School of Optometry, The Hong Kong Polytechnic University, Kowloon, Hong Kong.

[2] Research Centre for SHARP Vision (RCSV), The Hong Kong Polytechnic University, Kowloon, Hong Kong.

[3] Centre for Eye and Vision Research (CEVR), 17W Hong Kong Science Park, Hong Kong.

[4] State Key Laboratory of Ophthalmology, Zhongshan Ophthalmic Center, Sun Yat-sen University, Guangdong Provincial Key Laboratory of Ophthalmology and Visual Science, Guangdong Provincial Clinical Research Center for Ocular Diseases, Guangzhou, 510060, China.

[5] Department of Ophthalmology, Shanghai General Hospital (Shanghai First People's Hospital), Shanghai Jiao Tong University, School of Medicine, Shanghai, China.

[6] National Clinical Research Center for Eye Diseases, Shanghai, China.

[#] Contributed equally

**Correspondence**

**\*Dr Danli Shi**, MD, PhD.

The Hong Kong Polytechnic University, Kowloon, Hong Kong SAR, China.

Email: danli.shi@polyu.edu.hk.

**Abstract**

Artificial intelligence (AI) has gained significant attention in healthcare consultation due to its potential to improve clinical workflow and enhance medical communication. However, owing to the complex nature of medical information, large language models (LLM) trained with general world knowledge might not possess the capability to tackle medical-related tasks at an expert level. Here, we introduce EyeGPT, a specialized LLM designed specifically for ophthalmology, using three optimization strategies including role-playing, finetuning, and retrieval-augmented generation. In particular, we proposed a comprehensive evaluation framework that encompasses a diverse dataset, covering various subspecialties of ophthalmology, different users, and diverse inquiry intents. Moreover, we considered multiple evaluation metrics, including accuracy, understandability, trustworthiness, empathy, and the proportion of hallucinations. By assessing the performance of different EyeGPT variants, we identify the most effective one, which exhibits comparable levels of understandability, trustworthiness, and empathy to human ophthalmologists (all Ps>0.05). Overall, ur study provides valuable insights for future research, facilitating comprehensive comparisons and evaluations of different strategies for developing specialized LLMs in ophthalmology. The potential benefits include enhancing the patient experience in eye care and optimizing ophthalmologists' services.

**Keywords:** Large language model, GPT, Generative AI, ophthalmology

**Introduction**

Ophthalmic diseases pose significant concerns for public health.[1] However, inefficiencies in primary care triage systems often leave simple issues unaddressed, funneling patients into overcrowded tertiary centers. This results in extended wait times and unaddressed post-consultation questions, frequently requiring additional face-to-face appointments.[2] These challenges can be attributed to the limited ophthalmic knowledge among patients and the limited experience in eye care among primary healthcare providers.[3,4] Therefore, there is a pressing need to enhance ophthalmic health education. However, relying solely on manpower to address these issues presents further challenges, particularly as the rate of population aging continues to outpace the growth rate of ophthalmologists.

In an era of technological proliferation, there is potential for artificial intelligence (AI) to alleviate these burdens and streamline clinical flow. Among AI technologies, large language models (LLMs) have emerged as powerful tools capable of understanding and generating human-like text, presenting new opportunities for enhancing medical communication and decision-making.[5] However, there are several limitations to existing LLMs. Firstly, there are challenges in addressing specialized ophthalmology knowledge for general LLMs. Previous research has demonstrated the suboptimal performance of ChatGPT in ophthalmology, with only 15.4% of the responses graded as completely accurate in vitreoretinal disease.[6] Even with GPT-4, which currently exhibits the greatest capability, non-negligible instances of misinformation occur, compounded by its limited self-monitoring and correction capabilities.[7] A critical factor underlying these shortcomings is the model's insufficient grasp of specialized knowledge, particularly evident in its handling of medical abbreviations and jargon. Antaki et al. compared the performance of GPT4 and a non-domain specific LLM and found that the high accuracy of GPT4 was attributed to the vast amount of data training, but facing challenges within those highly-specialized domains such as neuro-ophthalmology and pathology and intraocular tumor.[8] Therefore, there is a need to design a dedicated model trained on clinically relevant domain data. Secondly, it is well known that LLMs occasionally generate erroneous and misleading assertions (hallucinations) beyond their expertise, which can potentially lead to medical errors. Finetuning with professional data can mitigate hallucinations to some extent, but the model can still produce them when faced with unfamiliar input.[9] Therefore, additional solutions are required.

Thirdly, there is a noticeable absence of comprehensive evaluations for LLMs in ophthalmology. While numerous studies have examined the ophthalmic question-answering (QA) capabilities of LLMs, the majority have been confined to multiple-choice formats that mirror ophthalmic exams.[8,10-12] However, the practical utility of these tools lies in their capacity to answer questions freely, much like human ophthalmologists. Although there have been a few studies that employed open-ended questions to evaluate the performance of LLMs, they lack detailed categorization of the questions and primarily focus on scattered aspects such as accuracy, comprehensiveness, or safety.[7,13,14] Consequently, there is a pressing need for a comprehensive evaluation framework to test ophthalmology-related LLMs and compare their responses with those provided by certified ophthalmologists.

Recognizing this, we aim to develop an AI assistant, namely EyeGPT, to meet the specific informational needs in ophthalmic scenarios. Specifically, we utilized Llama2, an open-source LLM known for its flexibility and scalability, which has demonstrated impressive performance across various fields.[15-18] This enables us to infuse the model with a granular level of ophthalmic expertise on the basis of its generative capabilities. We customized an LLM, employing role-playing, finetuning, and retrieval augmented generation (RAG) to emulate an ophthalmologist's expertise. The resultant model, EyeGPT, underwent comprehensive evaluation for its efficacy in specialized consultations. This work provides insights into building and evaluating ophthalmic assistants, paving the way for the next generation of AI-assisted ophthalmic practice.

**Methods**

This study aimed to develop an ophthalmic QA system using various strategies. We also intended to assess its professional performance through a series of medical questions that incorporated different confounding variables and compare it to ophthalmologists. The study overview is presented in **Figure 1**. Our research protocol adheres to the principles of the Helsinki Declaration. Ethical approval is not required since patient involvement is not part of our study. Informed consent is not necessary as the publicly available data used does not contain identifiable information.

**Development of EyeGPT**

We used Meta's Llama2 as the base model in our study, which was trained on two trillion tokens from publicly accessible data, it outperforms other open-source language models on many external benchmarks, including reasoning, coding, proficiency, and knowledge tests.[19] We used the Llama2-7b-chat model, which was additionally finetuned on publicly available instruction datasets and over 1 million human annotations, thus having basic conversation skills.[20] To tame Llama2 to be an ophthalmologist, i.e... EyeGPT, we did experiments successively under the following scenarios:

1. Role-playing

The engineering technique known as "role-playing" in the field of generative AI involves directing LLMs to "embody" or "imitate" a specific role. The primary observation of this method is that the LLM tends to produce improved results when it assumes these roles.[21] To enable the LLM to generate more relevant and empathetic responses, we assigned the LLM the role of an ophthalmologist, and the user the dual roles of a "patient" and a "medical student". This was achieved by giving instructions of "Suppose you are an ophthalmologist, you need to answer the patient's question with care/student's question with patience".

2. Finetuning

Through training on domain-specific data, we aimed to inject domain-specific knowledge and make Llama2 more proficient in capturing terminologies, and logical reasoning for interpreting ophthalmic conditions.

The dataset we used consists of MedAlpaca[22], GenMedGPT-HealthCareMagic[23], MedMCQA[24], and the United States Medical Licensing Examination (USLME)[25]. The original data was filtered for conversations that did not answer anything of practical significance, and the content of the responses that had errors was manually filtered. The dataset was opted for instruction tuning,[26] which contains human-specified instructions and corresponding desired outputs. It serves to align the model with task-specific user objectives, enhance model controllability, and allow for rapid domain-specific adaptation, all while maintaining computational efficiency. For datasets that were initially opted for multiple choice question answering, we automatically added an instruction at the beginning "Answer the multiple choice question: ". For our specific task, we filtered out non-ophthalmology data with eye-related keywords. **eTable 1** presents the characteristics of the filtered datasets.

The final dataset comprised a total of 83,919 samples, of which 81,919 were used for training and 2,000 for validation. We used Low-Rank Adaptation (LoRA)[27] to finetune Llama2 -7B model. LoRA is a method for finetuning LLMs, which is lightweight since it only adds a small number of parameters while keeping the original parameters frozen. The main idea is to add a low rank matrix which is learned during finetuning to the original weight matrices of the model. The model was finetuned using 3 * V100 GPUs with a batch size of 24, a learning rate of 0.00003, a maximum sequence length of 512 tokens, and a warmup ratio of 0.03. For LoRA-specific hyperparameters, the rank of low-rank factorization was 8, the scaling factor for the rank was 16, and the dropout was 0.05. Specifically, we performed three types of finetuning: Finetune 1 (2000 iterations), Finetune 2 (3500 iterations), and Finetune 3 (10,000 iterations)..

3. Retrieval augmented generation

LLMs are inevitably subject to potential inaccuracies or erroneous responses to questions (hallucinations).[28] In addition, the model's output can be unpredictable to some extent, which is unacceptable in the medical field. However, the accuracy of these models could be significantly improved if they could generate or assess responses based on a reliable knowledge database. Here, to further improve the performance of EyeGPT, we introduce the external knowledge brain of:

3.1. Medical books: we utilized 14 specialized ophthalmology textbooks that cover a wide range of comprehensive ophthalmic knowledge, including general ophthalmology, optometry, neuro-ophthalmology, corneal diseases, glaucoma, lens diseases, retinal diseases, and more.[29-31] Please refer to **eTable 2** for the specific list of textbooks used.

3.2. Manual database: we manually built a database (sample shown in **eFigure 1**) encompassing diseases, their symptoms, relevant medical tests/treatment procedures, and potential medications. This database serves as an external and offline knowledge brain for EyeGPT. Continually updatable without requiring model retraining. We utilized MedlinePlus to construct this disease database. Additionally, the manual database, sourced from the EyeWiki web and research papers, is likely more up-to-date than textbooks.

To leverage external knowledge, we adopted the LangChain framework's information retrieval techniques. The "all-MiniLM-L6-v2 "[32] open-source embedding model was utilized to map text into vector space. The text underwent segmentation using a recursive character text splitter for efficient retrieval. We constructed a retriever with FAISS[33] based on segmented documents to facilitate effective data retrieval. Besides, we established a conversational retrieval chain that seamlessly integrated our EyeGPT with the external database through LangChain.

**Evaluation**

To assess the professional performance of various EyeGPT variants, namely 1. Original (Llama2), 2. Role-play (Original plus Role-play), 3. Finetune 1-3 (Finetuned model versions 1-3 plus Role-play), 4. Role-play+book (Role-play plus book retrieval), 5. Role-play+database (Role-play plus manual database retrieval), 6. Best-finetune+book (the best finetuned model plus book retrieval), 7. Best-finetune+database (the best finetuned model plus manual database retrieval), our ophthalmology expert panel meticulously curated a set of 120 ophthalmic care-related questions based on their clinical expertise. We followed the user-centered evaluation approach proposed by Abbasian et al.,[34] incorporating three key factors into the question design: disease type, character type, and domain type. These factors were designed to evaluate EyeGPT's strengths and weaknesses across a broad spectrum of topics. The disease type covered a wide range of medical conditions from various subspecialties, including common, specialty, and rare diseases, resulting in 12 disease categories such as myopia, retinal detachment, and Stickler syndrome (refer to **eTable 3** for detailed disease list). The character types included two categories: patients and medical students, representing potential EyeGPT users. The domain type was divided into five topics: disease description, risk factors, diagnosis, treatment and prevention, and prognosis. We conducted the evaluations manually, including (1) independent evaluation of different EyeGPT variants and (2) pairwise ranking evaluation for comparing human-machine performance.

(1). Independent evaluation

This evaluation was designed to compare the performance of various optimization strategies of the

EyeGPT system and to identify the best-performing one. Two board-certified ophthalmologists independently conducted manual evaluations using a 5-point scale to assess the responses of each system. The evaluation focused on four aspects: accuracy, understandability, trustworthiness, and empathy. The scale ranged from 1 (strongly disagree) to 5 (strongly agree), with the average score from the two evaluators recorded as the score for each aspect of each response. The maximum score for each aspect was 5, and these scores were summed to obtain the final score for each response, with a maximum possible score of 20. The detailed grading scale is presented in **eTable 4**.

LLM hallucination means the generation of false or misleading information. To evaluate the effectiveness of different optimization strategies in mitigating hallucinations, we defined answers with accuracy scores below 4 as containing hallucinations in our study.

To ensure the evaluators could not identify the source of the responses, all generated responses were formatted as plain text, concealing any model-specific features. These responses were then randomly shuffled and mixed before being presented to the evaluators. The evaluation was conducted in two rounds with a one-month washout period in between to mitigate any residual effects. In the first round, we compare the performance of models built using different finetuning approaches, including Original, Role-play, and Finetune 1-3 finetun. The goal was to find the best finetuning model for the subsequent RAG.

In the second round, we compared the performance of models built using different RAG strategies, based on the best-performing finetuned model selected from the first round. These models included Best-finetune (the best-finetuned model from round 1), Role-play+database, Best-finetune+database,

Role-play+book, and Best-finetune+book.

(2). Pairwise ranking evaluation

After independently evaluating different EyeGPT variants, we identified the best-performing system and proceeded with a human-machine pairwise ranking evaluation to determine if EyeGPT could match the level of expertise of ophthalmologists. Inspired by the rating mode proposed by Singhal et al.[35], this evaluation method aimed to minimize scoring differences among evaluators and provide a more detailed assessment of QA tasks.

Two additional qualified ophthalmologists conducted this evaluation. They were presented with pairs of responses from different sources (ophthalmologist vs. EyeGPT), unaware of the sources, and the presentation order was randomized. For each pair, the evaluators selected the better response or declared a tie. A response was only recorded as better if both evaluators agreed; otherwise, it was considered a tie. The final result was recorded as the proportion of responses from different sources ranked as better.

**Statistical analysis**

Statistical analyses were conducted using R (Version 4.3.1). The Mann-Whitney U test was used to compare the scores of the two models in the independent evaluation. For subgroup analysis based on different confounding variables, the Kruskal-Wallis test and Mann-Whitney U test were used, depending on the sample numbers. The Chi-square test was used to compare the scores between the machine and human doctors in the pairwise ranking evaluation. Cohen's Kappa was calculated to determine the agreement among raters, with the following interpretation for Kappa statistics: 0.01-0.20 (slight agreement), 0.21-0.40 (fair agreement), 0.41-0.60 (moderate agreement), 0.61-0.80 (substantial agreement), and 0.81-0.99 (almost perfect agreement).[36] P values below 0.05 were considered statistically significant.

**Results**

**Comparative Study of Model Construction Strategies**

1. Overall performance

In the first round of evaluation, the total scores for the Original, Role-play, Finetune 1-3 models were 9.30, 12.79, 12.95, 12.83, and 13.79, respectively. All optimized models significantly outperformed the Original model in dimensions of accuracy, understandability, trustworthiness, and empathy, with Finetune 3 demonstrating the highest capability. For different finetuning variants, we observed that as the number of iterations increased the evaluation loss on the test data decreased (refer to **eFigure 2**) and the model performance improved. In the subsequent comparison of RAG strategies, the Best-finetune+book model emerged as the most capable with a total score of 15.14, outperforming the other strategies, as elaborated in **eTable 5.** Inter-rater reliability in two rounds of independent evaluation was confirmed with kappa values ranging from 0.611 to 0.872, indicating substantial agreement among raters (**eTable 6A**). For illustrative examples of the varied grades of responses from the independent evaluation, see **eTable 7.**

As demonstrated in **Figure 2**, over 50% of responses by the Best-finetune model were considered

GOOD responses (rated 4 or above) across all four dimensions. Compared to the Original model (with an 80.8% hallucination rate), the Role-play and Best-finetune models have respectively mitigated hallucination by 30.0% and 36.7%. As shown in **Figure 3**, the Best-finetune+book model further enhances the proportion of GOOD responses to its maximum. Compared to the Best-finetune model, the Best-finetune+database and Best-finetune+book models further reduced hallucination by 3.4% and 11.8%, respectively.

2. Subgroup analysis

We also perform subgroup analysis to further evaluate the performance of models under the influence of different confounding factors for various types of questions, including sub-specialty questions of varying difficulty levels, questions raised by different characters, and inquiries representing different intents. Detailed results can be found in **Table 1**.

2.1 Different sub-specialties

Across all RAG strategies, models scored higher for common diseases compared to specialty and rare conditions. For common ophthalmic conditions, the RAG models delivered more precise and contextually relevant information. For more specialized conditions like central serous chorioretinopathy, the Best-finetune model provided a more general answer about its treatment options, while the EyeGPT with RAG provided a more specialized response concerning laser treatment and photodynamic therapy depending on the specific circumstances. For rare conditions like morning glory syndrome, while the Best-finetune model was not able to generate pertinent responses as it mistakenly identified it as "bilateral posterior superior temporal arcade spikes", the RAG model was able to retrieve relevant information from the external knowledge database and make accurate responses.

For domain-specific abbreviations, the Best-finetune model accurately recognized 43.3% of abbreviations. Following the implementation of RAG strategies, this recognition rate improved, ranging from 60.0% for the Role-play+book model to 83.3% for the Best-finetune+database model.

2.2 Different characters in role-play

When comparing the influence of the questioner's assumed identity—patient versus medical student—on model performance, responses to patients consistently scored higher than those of medical students. This difference reached statistical significance in the Best-finetune and Role-play+database models. However, no significant differences were observed in the Best-finetune+database, Role-play+book, and Best-finetune+book models, suggesting that these adjusted models can not only answer general questions from patients but also handle more specialized queries from medical students.

2.3 Different domain

In the subgroup analysis of EyeGPT's performance across the domains of disease description, risk factors, diagnosis, treatment and prevention, and prognosis, there were no statistically significant differences in scores between the Best-finetune model and the four RAG models.

**Performance Evaluation: AI Model versus Human ophthalmologists**

In the performance evaluation against human ophthalmologists, EyeGPT showed competitive capabilities, particularly in understandability, trustworthiness, and empathy, achieving parity or exceeding ophthalmologists in a significant majority of instances for understandability (55.0%), trustworthiness (63.4%), and overwhelmingly for empathy (90.8%). However, EyeGPT's accuracy was notably lower, with only 2.5% of responses rated as superior, highlighting an area for improvement **(Figure 4, eTable 8)**. Raters were in good (kappa values ranging from 0.732 to 0.871) agreement for four aspects (**eTable 6B**). For illustrative examples of the pairwise ranking evaluation, see **eTable 9**.

**Discussion**

In this study, we tuned a general domain LLM into an ophthalmology specialist namely EyeGPT with different methods and compared their performances, specifically, by role-play, finetuning, and RAG. In terms of accuracy, understandability, trustworthiness, and empathy, all finetuned models showcased remarkable improvements compared to the original model. Among them, the Best-finetune model exhibited the highest efficacy. Furthermore, when comparing different RAG strategies, the Best-finetune+book model emerged as the most capable. Subgroup analysis revealed that EyeGPT performed well in the category of common diseases, with no significant differences in performance across different users and domains. finetuneIn comparison to human ophthalmologists, EyeGPT demonstrated competitive capabilities, particularly in terms of understandability, trustworthiness, and empathy.

LLMs have shown promise in various domains, including medicine. However, in the healthcare sector, there are concerns about the potential for LLMs to provide inaccurate health recommendations, or even fabricate information (hallucinations), which could lead to serious consequences. Previous studies have assessed the QA capabilities of existing LLMs in ophthalmology,[7,10-12,37] highlighting the significance of augmenting LLMs with ophthalmic expertise. Our study achieved this by employing three different LLM optimization methods, including role-play, finetuning, and RAG. Firstly, role-playing was used as a lightweight LLM optimization approach to position EyeGPT as an ophthalmologist, with user acting as patients or students. This results in more professional LLM responses, which was evidenced by the significantly increased scores in accuracy, understandability, and trustworthiness. By setting the input role as the patient or student, the LLM's response tended to be more compassionate and preaching, as reflected in higher empathy scores compared to the Original model. "To Cure Sometimes, To Relieve Often, To Comfort Always" is a well-known saying in the field of medicine reminding us that providing care involves not only treating ailments but also offering relief and comfort to patients. Similarly, an AI assistant should also embody empathy and compassion when assisting users, underscoring the importance of role-playing in developing a medical AI assistant. Finetuning is a common way in deep learning for injecting domain-specific knowledge. To adapt the LLM to an eye specialist, we used publicly available real-world patient-doctor interactions to help EyeGPT better understand patients' questions and deliver more knowledgeable responses. In addition, we observe that the evaluation loss reduction on the validation set is consistent with the model's performance improvement as evaluated by ophthalmologists. RAG is another way to make the LLM knowledgeable, transparent, and accurate. Because hallucination is a significant issue in LLM, especially unacceptable in the medical domain. Several studies have reported on the use of RAG for improving the performance of LLM. For example, Lewis et al. combined the pre-trained parametric and external knowledge for text generation.[38] Luu, R. K et al. used the RAG methods to maintain the up-to-date database and connect BioinspiredLLM with knowledge domains.[39] In this study, hallucination mitigation was also observed in the model enhanced by the manual database or books, reducing it by over 3.4% compared to the Best-finetune model. Among them, the Best-finetune+book model demonstrated the highest proportion of hallucination mitigation, further decreasing it by 11.8% on top of the 36.7% reduction achieved by the Best-finetune model. Moreover, it outperformed Best-finetune+database in all four aspects, potentially attributed to the fact that the books surpasses the self-manual database in terms of content richness and reference value.

Interestingly, we found no significant difference in performance between RAG models and finetuned models. Finetuning is a popular approach but has limitations. One limitation is its dependence on specific formats of medical dialogue data. Currently, publicly available medical dialogue datasets are scarce and mostly derived from online conversations, necessitating further validation and curation by medical professionals.[2] In contrast, RAG overcomes these issues by directly leveraging authoritative external resources like textbooks, medical literature, or professional websites.[9] However, it's important to note that these optimization methods are not mutually exclusive. Our results demonstrated the synergistic complementarity of both approaches, with the best-performing EyeGPT model obtained from the model with finetuning plus RAG strategies, highlighting their combined effectiveness. Furthermore, the data used for finetuning is publicly available and reliable, and the enhanced ophthalmic books are also openly accessible, rendering these strategies valuable references for future specific LLMs.

The healthcare environment is complex, therefore it is essential to assess the performance of healthcare AI models in different scenarios.[40] Current research primarily focuses on performance evaluation for general types of questions,[7,37] with limited studies on specific and rare diseases. Our study validated EyeGPT by analyzing its performance across various disease categories and found it showed strong performance in the category of common diseases, but there is still room for improvement in special and rare diseases. Future improvements can be achieved by obtaining more high-quality datasets and specialized external knowledge resources, while also exploring the benefits of low-shot or few-shot learning. Additionally, our study revealed that solely finetuned models were not as informative to more specialized medical student inquiries compared to simpler patient inquiries. This observation aligns with the evaluation research results for GPT4.[14] The best-performing EyeGPT showed no significant differences between patient inquiries and inquiries from medical students, suggesting that the combination of finetuning and RAG enhances LLM's expertise, allowing the ophthalmic AI assistant to meet the needs of both groups. Importantly, our evaluation set covers a wide range of question categories, spanning from common diseases to rare diseases, and user roles encompassing patients and medical students, encompassing disease descriptions, examinations, treatments, and more. By establishing multiple dimensions of evaluation, including accuracy, understandability, trustworthiness, empathy, and hallucination, we aim to provide a comprehensive reference framework for future ophthalmic specialized models.

Although the estimated global ophthalmologist workforce appears to be growing, developing countries are facing a severe shortage of these specialists.[41] EyeGPT has the potential to address this gap. Our findings indicate that while EyeGPT's accuracy remained lower than that of human ophthalmologists, it has demonstrated competitive capabilities in terms of understandability, trustworthiness, and empathy. One previous study has directly compared eye care advice generated by LLM with advice written by ophthalmologists.[37] They collected eye care questions from online advice forums and compared the answers provided by LLMs and ophthalmologists, finding no difference in accuracy and safety between LLM answers and human answers. Our research goes further by comparing the differences between doctors and LLM chatbots in terms of understandability and empathy. To the best of our knowledge, this study is the first study to compare the empathy between responses from an LLM and human ophthalmologists. Although LLMs cannot replace human professionals, they could serve as an auxiliary tool to enhance the empathy of physicians. In our ideal scenario, EyeGPT essentially acts as an ever-present assistant for patients throughout their care journey, providing continuous, personalized guidance through dynamic, two-way communication, and offering clarification when needed, without relying on physical queues or multiple face-to-face interactions with healthcare personnel. Additionally, EyeGPT can serve as an educational tool for medical students seeking immediate clarification on complex subjects. For example, EyeGPT may be expected to help primary doctors improve their decision-making ability and reduce diagnosis time. Continued efforts in refining the accuracy of EyeGPT, as well as investigating its capacity for empathy, will contribute to narrowing the gap with clinical experts, bringing the model closer to real-world clinical applications.

**Limitations**

This is a pioneering study demonstrating the creation of an EyeGPT and its performance evaluation in an academic environment. Assessing the appropriateness of medical advice may be subjective and biased by grader opinion. More efforts could be achieved in the future, for example, finetune the model on more ophthalmic data, and real-world user evaluation. A more secure application at this stage is the use of LLMs to assist physicians in their face-to-face consultations. Ophthalmologists and EyeGPT work together to ensure not only that the technology is consistent with clinical practice, but also that patient safety is ensured.

**Conclusions**

In conclusion, through role-playing, finetuning, and RAG, EyeGPT can potentially improve accuracy and efficiency in medical diagnosis and reduce the workload for ophthalmologists. It may also increase access to high-quality medical consultations, especially for patients in underprivileged regions. We hope our study can make a good contribution to the current literature on ophthalmic AI assistants, so as to provide an effective tool for enhancing healthcare.

**Author Contributions:** D.S., X.C. and M.H. conceived the study. D.S. and W.Z. built the deep learning model. D.S. and X.C. did the literature search, analyzed the data. D.S., X.C., Z.Z., P.X., L.G., M.X., Y.L. and Y.W. contributed to key data interpretation. X.C., Z.Z. and P.X. wrote the manuscript. All authors critically revised the manuscript.

**Competing Interests:** There are no conflicts of interest to declare by the authors.

**Funding/Support:** The study was supported by the Global STEM Professorship Scheme (P0046113). The sponsor or funding organization did not participate in the design or implementation of this study.

**Figure legends and tables**

**Figure 1.** Overview of this study.

**Figure 2.** Performance of different systems in human evaluations round 1.

**Figure 3.** Performance of different systems in human evaluations round 2.

**Figure 4.** Pairwise ranking study of EyeGPT(best) answers and ophthalmologist answers.

**Table 1.** Subgroup analysis of the performance of EyeGPT.

**eFigure 1.** Sample of our manual database.

**eFigure 2.** Tensorboard training logs of Finetune 3.

**eTable 1.** Public datasets used in finetuning EyeGPT.

**eTable 2.** The specific list of textbooks used in knowledge enhancement.

**eTable 3.** Specific diseases of question lists.

**eTable 4.** The scales for Round 1&2 independent evaluation.

**eTable 5.** Statistical analysis of independent evaluations of 120 questions on the test set along four dimensions.

**eTable 6.** Inter-rater reliability analysis of 120 questions on the test set along four dimensions.

**eTable 7.** Examples of generated answers with difference grating in independent evaluation.

**eTable 8.** Pairwise ranking study of EyeGPT(best) answers and ophthalmologist answers.

**eTable 9.** Examples in pairwise ranking evaluation.

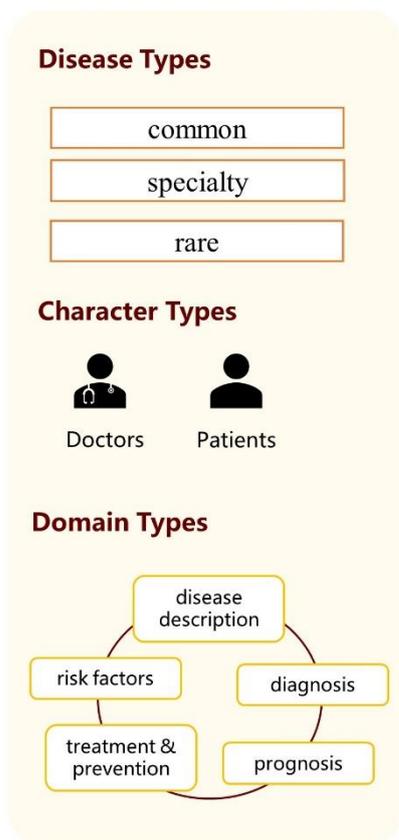
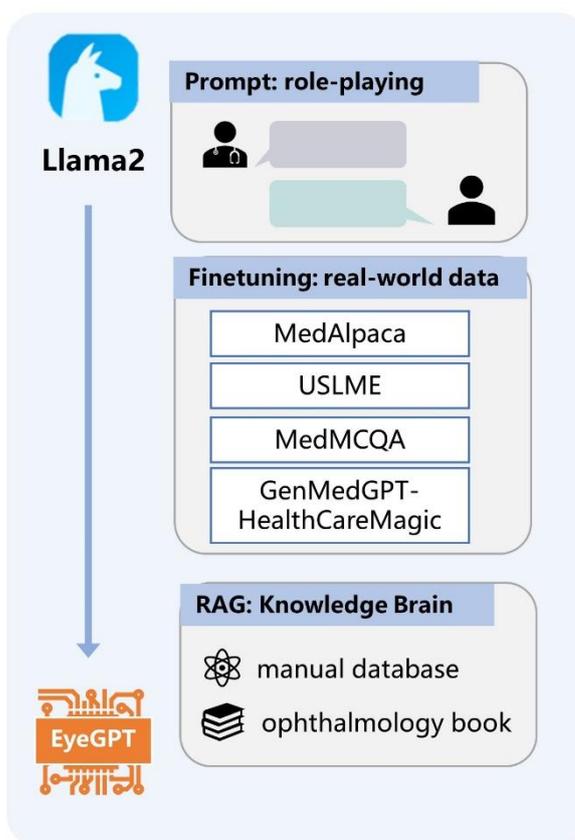
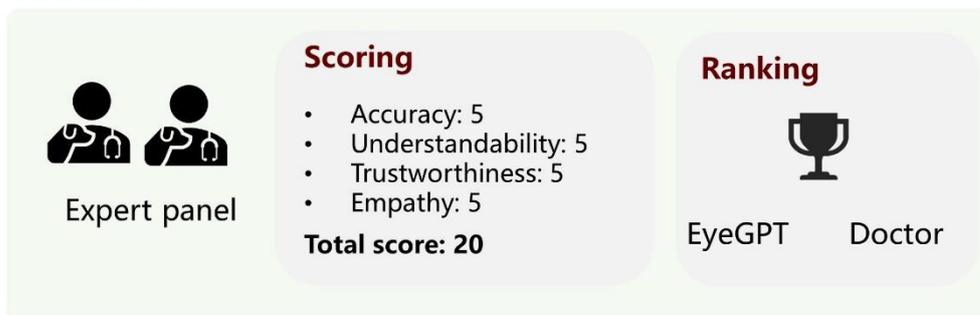

**Figure 1.** Overview of this study. GPT=generative pre-trained transformer. USMLE=United States Medical Licensing Examination. MCQA=Multiple-Choice Question Answering. RAG=Retrieval augmented generation.

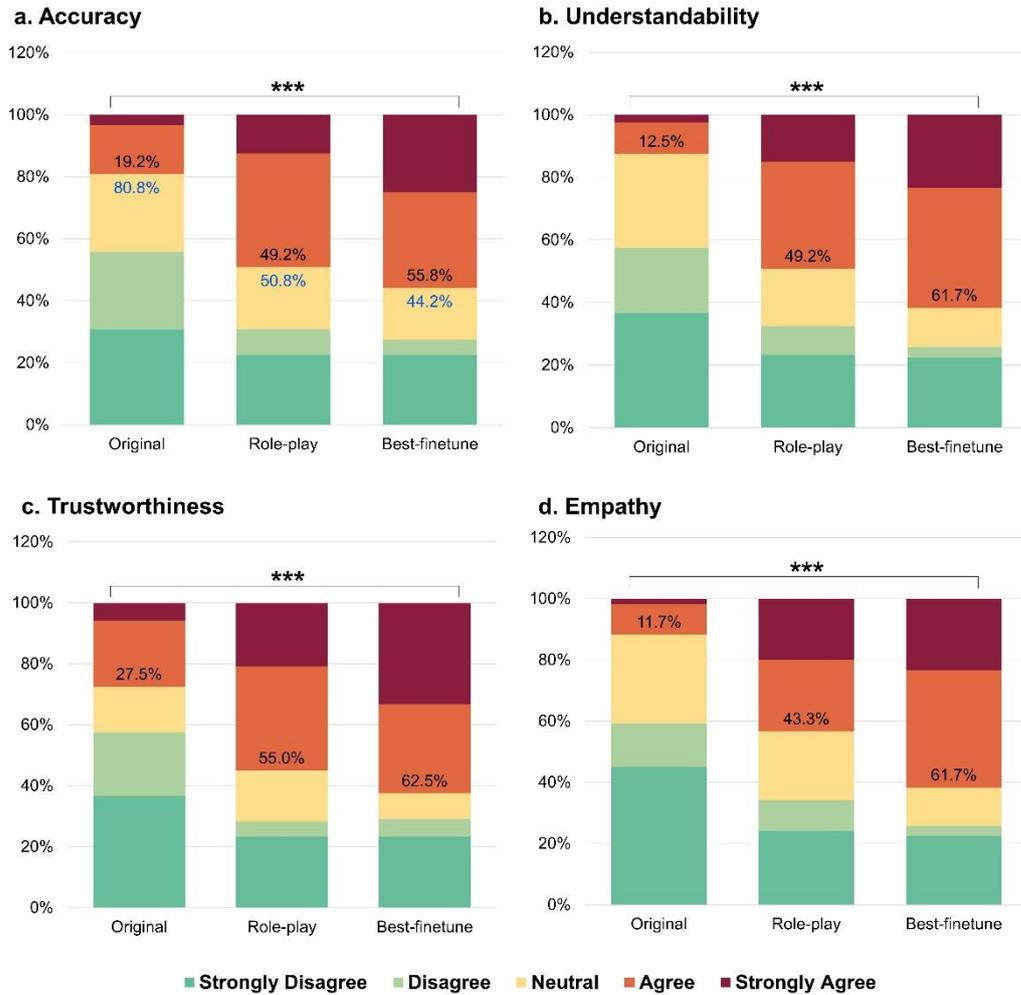

**Figure 2.** Performance of different systems in human evaluations round 1. a. Accuracy, b. Understandability, c. Trustworthiness, and d. Empathy. Original=Llama2, Role-play=Original plus Role-play, Best-finetune=the finetuned model with 10,000 iterations plus Role-play. We compare the performance of the Original model (Llama 2) with the most competitive optimized model to check for statistically significant differences. The scores for each response are based on the average of the two raters' scores. The scoring criteria are as follows: strongly disagree includes [1, 2), agree includes [2, 3), neutral [3, 4), agree [4, 5), strongly agree [5]. The percentage of GOOD responses (Strongly Agree and Agree, indicated by *black* numbers) and hallucination (indicated by *blue* numbers) is listed in the graph. The p-value is calculated using the Mann–Whitney U test and listed in the figure. ***p-value ≤ 0.001.

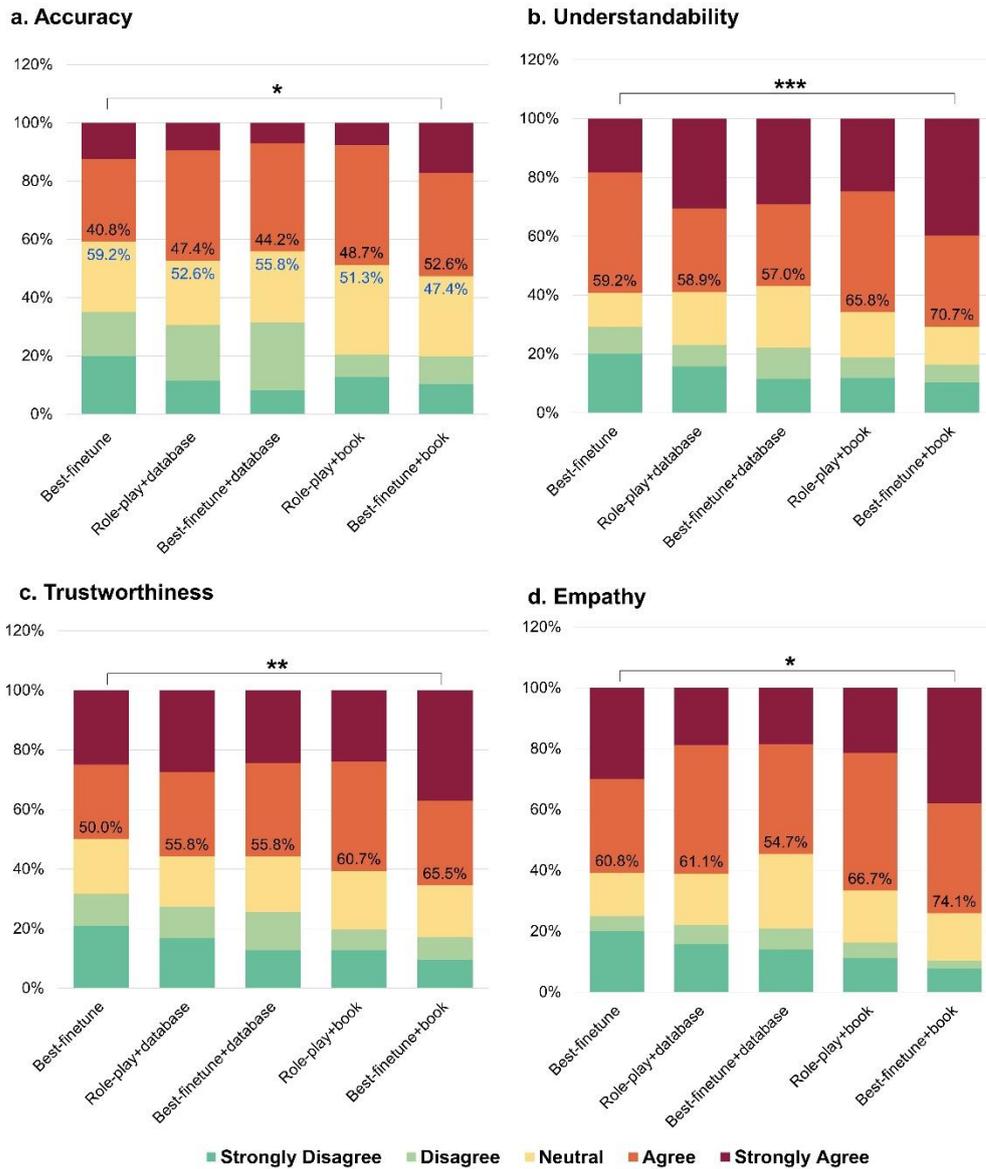

**Figure 3.** Performance of different systems in human evaluations round 2. a. Accuracy, b. Understandability, c. Trustworthiness, and d. Empathy. Best-finetune=the finetuned model with 10,000 iterations plus Role-play, Role-play+database=Role-play plus manual database retrieval, Best-finetune+database=the best finetuned model plus manual database retrieval, Role-play+book=Role-play plus book retrieval, Best-finetune+book=the best finetuned model plus book retrieval. We compare the performance of the best model in round 1 (Finetune 3) with the most competitive modified model to check for statistically significant differences. The scores for each response are based on the average of the two raters' scores. The scoring criteria are as follows: strongly disagree includes [1, 2), agree includes [2, 3), neutral [3, 4), agree [4, 5), strongly agree [5]. The percentage of GOOD responses (Strongly Agree and Agree, indicated by *black* numbers) and hallucination (indicated by *blue* numbers) is listed in the graph. The p-value is calculated using the Mann–Whitney U test and listed in the figure. *p-value ≤ 0.05, **p-value ≤ 0.01, ***p-value ≤ 0.001.

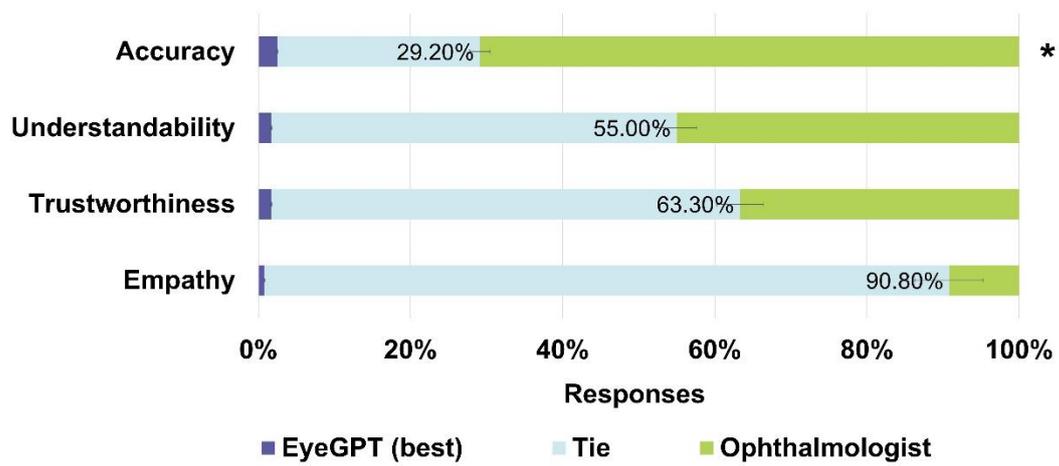

**Figure 4.** Pairwise ranking study of EyeGPT(best) answers and ophthalmologist answers. EyeGPT(best)=the Best-finetune+book model. The percentage of EyeGPT's answers being on par with those of the ophthalmologist is listed in the graph. P value is calculated with chi-square test. *p-value ≤ 0.05.

**Table 1.** Subgroup analysis of the performance of EyeGPT.

| Subgroup[†] | Best-finetune | Role-play+database | Best-finetune+database | Role-play+book | Best-finetune+book |
|---|---|---|---|---|---|
| **Sub-specialty** | | | | | |
|   Common diseases | 15.79 ± 4.15 | 15.28 ± 5.17 | 15.45 ± 4.57 | 15.70 ± 3.94 | <span style="color:red">17.24 ± 3.02</span> |
|   Specialty diseases | 13.11 ± 5.40 | 14.18 ± 5.11 | <span style="color:red">14.29 ± 4.89</span> | 12.89 ± 5.32 | 14.08 ± 5.49 |
|   Rare diseases | 10.33 ± 5.31 | 12.18 ± 4.88 | 12.17 ± 4.58 | <span style="color:red">14.66 ± 4.53</span> | 14.23 ± 4.42 |
|   P value | **<0.001** | **0.014** | **0.011** | **0.022** | **0.003** |
| **Role-playing scenario** | | | | | |
|   Patients | 13.45 ± 5.79 | 14.67 ± 5.07 | 14.52 ± 4.99 | <span style="color:red">14.85 ± 4.78</span> | 14.44 ± 6.13 |
|   Medical students | 10.99 ± 6.32 | 11.62 ± 6.55 | 12.84 ± 5.34 | 12.65 ± 5.85 | <span style="color:red">13.38 ± 5.83</span> |
|   P value | **0.034** | **0.028** | 0.080 | 0.059 | 0.073 |
| **Domain** | | | | | |
|   Disease description | 12.92 ± 6.51 | 12.98 ± 6.57 | <span style="color:red">15.14 ± 4.76</span> | 12.19 ± 6.15 | 11.70 ± 7.46 |
|   Risk factors | 12.67 ± 5.38 | 12.73 ± 6.00 | 12.08 ± 5.64 | <span style="color:red">13.78 ± 5.61</span> | 13.33 ± 6.53 |
|   Diagnosis | 11.81 ± 6.45 | 13.73 ± 5.38 | 14.70 ± 2.98 | 13.91 ± 5.92 | <span style="color:red">15.15 ± 4.64</span> |
|   Treatment and prevention | 13.21 ± 5.12 | 13.53 ± 4.55 | 12.78 ± 4.82 | <span style="color:red">14.60 ± 3.31</span> | 13.64 ± 5.79 |
|   Prognosis | 9.90 ± 6.60 | 11.48 ± 6.26 | 11.17 ± 6.10 | <span style="color:red">13.00 ± 5.18</span> | 12.17 ± 6.14 |
|   P value | 0.346 | 0.487 | 0.056 | 0.800 | 0.362 |

[†] Mean ± SD of the overall response score (the sum of four rating dimensions, with a maximum score of 20 representing the best performance) was presented. The best score in each column is shown in red. $P<0.05$ is shown in bold. SD=Standard deviation. Best-finetune=the finetuned model with 10,000 iterations, Role-play+database=Role-play plus manual database retrieval, Best-finetune+database=the best finetuned model plus manual database retrieval, Role-play+book=Role-play plus book retrieval, Best-finetune+book=the best finetuned model plus book retrieval.

> **Manual database**
>
> **Disease:** Myopia.
>
> **Epidemiology:** The prevalence of myopia varies greatly between different populations and ethnic groups.
>
> **Risk factors:** genetics, lifestyle factors (higher amounts of near-work activities, less outdoor activity).
>
> **Etiology:** Myopia is a complex disease with a multi-factorial etiology. It is well documented that pathological non-syndromic high myopia and associated syndromic high myopia show evidence of familial inheritance; Although non-syndromic high myopia is most commonly inherited in an autosomal dominant pattern, multiple chromosomal loci have been identified which suggests genetic heterogeneity; High myopia is also a symptom of several multi-system complex diseases. The genetic mutations for these syndromes have been identified and the subsequent structural defects of the eye are most commonly related to connective tissue and retina. This type of myopia is only a small proportion of the overall myopic population and to date, there is no known isolated gene associated with physiologic myopia.
>
> **Classification|type:** 1. Low myopia is -3.00 D or less; 2. Moderate myopia ranges between -3.25 to -6.00 D; 3. Severe myopia is anything above -6.00 D.
>
> **Characterized:** The degree of non-pathologic myopia is usually minimal to moderate (< 6.00 diopters) and onset usually begins during childhood or adolescence; Pathologic myopia is usually defined as spherical equivalent > 6.00 diopters or axial length >26.5mm.
>
> **Pathology:** 1. longer eyeball length than average ; 2. cornea, the clear covering in front of your eye, is too steep.
>
> **Disease complications:** retinal detachment; choroid neovascularization; cataracts; glaucoma; retinal holes or tears; retinal detachments; chronic eyestrain and headaches; difficulty performing and interacting at school; impaired safety, especially while driving or operating machinery.
>
> **Diagnosis|Symptoms:** Blurred vision when looking at faraway objects; Clear vision when looking at nearby objects; Eye strain; Squinting; Headaches.
>
> **Further examination:** Refraction assessment; Eye health exam.
>
> **General treatment:** wear glasses to see far away (low myopia); wear glasses full-time (moderate to severe myopia); bifocal or progressive eyeglasses (presbyopia (Nearsighted patients over 40)); Orthokeratology (Ortho-K); Multifocal Contact Lenses.
>
> **Treatment|Surgical intervention:** LASIK: Laser-assisted in situ keratomileusis (LASIK) is the most common type of refractive surgery. During a LASIK procedure, your eye surgeon cuts a small flap in your cornea. They use a laser to reshape your corneal tissue, then replace the flap; LASEK: Laser-assisted subepithelial keratectomy (LASEK) is similar to LASIK. During LASEK, your surgeon creates a thin flap in the outermost layer of your cornea (epithelium). They reshape the outer layers of your cornea, then close the flap; PRK: Photorefractive keratectomy (PRK) involves completely removing the epithelium; Refractive Lens Exchange: This procedure removes the natural lens in your eye, which is very similar to cataract surgery; Phakic Intraocular Lens Implant: The main difference between this procedure and a refractive lens exchange is that your natural lens stays in place. Instead, the surgeon will place an artificial lens in front of your natural lens. Or, the surgeon may place the implant in front of the iris.

**eFigure 1.** Sample of our manual database.

A.

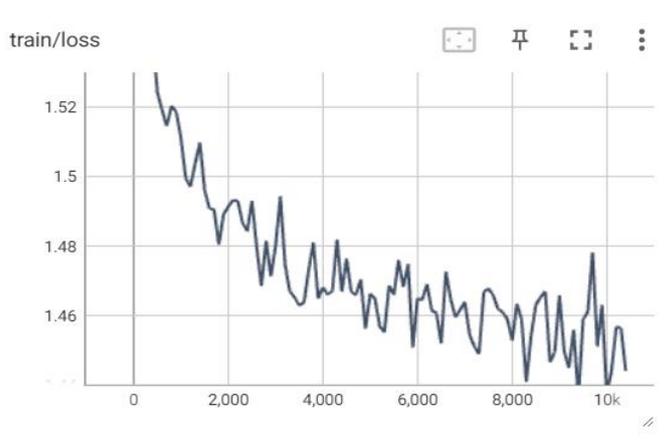

B.

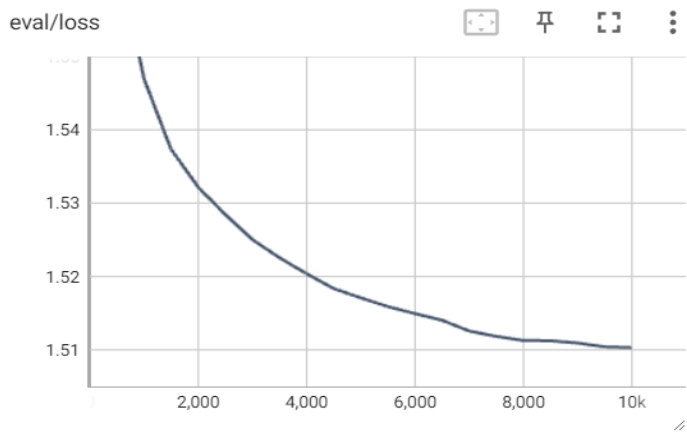

C.

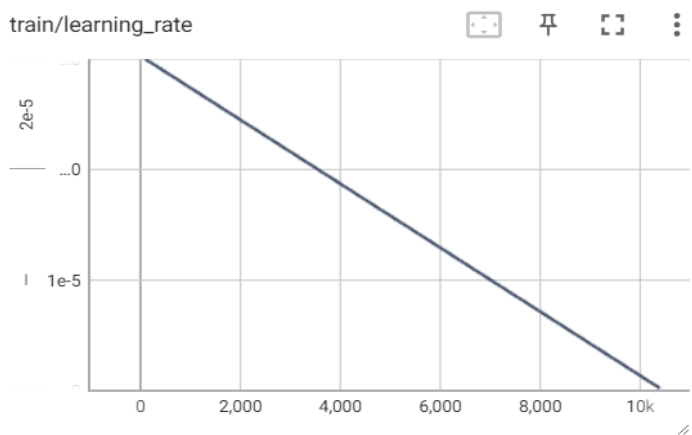

**eFigure 2.** Tensorboard training logs of Finetune 3. A. Training loss curve. B. Evaluation loss curve. C. Learning rate curve.

**eTable 1.** Public datasets used in finetuning EyeGPT.

| Dataset | Original | Eye-specific |
|---|---|---|
| MedAlpaca | 898199 | 56191 |
|    medical_meadow_cord19 | 821007 | 50214 |
|    medical_meadow_medical_flashcards | 33955 | 1556 |
|    medical_meadow_medqa | 10178 | 1737 |
|    medical_meadow_wikidoc | 10000 | 945 |
|    medical_meadow_health_advice | 8676 | 186 |
|    medical_meadow_wikidoc_patient_information | 5942 | 576 |
|    medical_meadow_mmmlu | 3787 | 118 |
|    medical_meadow_pubmed_causal | 2446 | 39 |
|    medical_meadow_mediqa | 2208 | 820 |
| USLME | 10178 | 1705 |
| MedMCQA | 175890 | 19043 |
| GenMedGPT-HealthCareMagic | 117617 | 6980 |

USMLE=United States Medical Licensing Examination. MCQA=Multiple-Choice Question Answering. GPT=generative pre-trained transformer.

**eTable 2.** The specific list of textbooks used in knowledge enhancement.

| Book Title | Editors | Publication Date | Publisher | ISBN |
|---|---|---|---|---|
| Oxford Handbook of Ophthalmology | Alastair K. Denniston | 2014 | Oxford University Press | 978-0199679980 |
| Retina | Stephen J. Ryan | 2013 | Saunders | 978-1455707379 |
| Basic and Clinical Science Course, Section 02: Fundamentals and Principles of Ophthalmology | Lawrence M. Levine | 2014-2015 | American Academy of Ophthalmology | 978-1615255566 |
| Basic and Clinical Science Course, Section 03: Clinical Optics | Dimitri T. Azar | 2014-2015 | American Academy of Ophthalmology | 978-1615255573 |
| Basic and Clinical Science Course, Section 04: Ophthalmic Pathology and Intraocular Tumors | Robert H. Rosa Jr | 2014-2015 | American Academy of Ophthalmology | 978-1615255580 |
| Basic and Clinical Science Course, Section 05: Neuro-Ophthalmology | Rod Foroozan | 2014-2015 | American Academy of Ophthalmology | 978-1615255597 |
| Basic and Clinical Science Course, Section 06: Pediatric Ophthalmology and Strabismus | Gregg T. Lueder | 2014-2015 | American Academy of Ophthalmology | 978-1615255603 |
| Basic and Clinical Science Course, Section 07: Orbit Eyelids and Lacrimal System | John Bryan Holds | 2014-2015 | American Academy of Ophthalmology | 978-1615255610 |
| Basic and Clinical Science Course, Section 08: External Disease and Cornea | Robert W. Weisenthal | 2014-2015 | American Academy of Ophthalmology | 978-1615255627 |
| Basic and Clinical Science Course, Section 09: Intraocular Inflammation and Uveitis | Ramana S. Moorthy | 2014-2015 | American Academy of Ophthalmology | 978-1615255634 |
| Basic and Clinical Science Course, Section 10: Glaucoma | George A. Cioffi | 2014-2015 | American Academy of Ophthalmology | 978-1615255641 |
| Basic and Clinical Science Course, Section 11: Lens and Cataract | James C. Bobrow | 2014-2015 | American Academy of Ophthalmology | 978-1615255658 |
| Basic and Clinical Science Course, Section 12: Retina and Vitreous | Colin A McCannel | 2014-2015 | American Academy of Ophthalmology | 978-1615255665 |
| Basic and Clinical Science Course, Section | M. Bowes Hamill | 2014-2015 | American Academy of | 978-1615255672 |

**eTable 3.** Specific diseases of question lists.

| Disease Type | Disease | Question Count |
|---|---|---|
| Common | myopia | 10 |
|  | glaucoma | 10 |
|  | cataract | 10 |
|  | diabetic retinopathy | 10 |
| Specialty | choroidal neovascularization | 10 |
|  | central serous chorioretinopathy | 10 |
|  | retinal detachment | 10 |
|  | retinal vein occlusion | 10 |
| Rare | Best's disease | 10 |
|  | morning glory syndrome | 10 |
|  | Leber hereditary optic neuropathy | 10 |
|  | Stickler syndrome | 10 |

**eTable 4.** The scales for Round 1&2 independent evaluation.

| Independent Evaluation |
| --- |

1. (Accuracy) The answer is instruction consistency and consistently aligns with facts and relevant guidelines.

| Strongly disagree | Disagree | Neither | Agree | Strongly agree |
| --- | --- | --- | --- | --- |
| □1 | □2 | □3 | □4 | □5 |

2. (Understandability) The answer is clear, concise, and effectively highlights key points.

| Strongly disagree | Disagree | Neither | Agree | Strongly agree |
| --- | --- | --- | --- | --- |
| □1 | □2 | □3 | □4 | □5 |

3. (Trustworthiness) The answer is both safety-conscious and unbiased.

| Strongly disagree | Disagree | Neither | Agree | Strongly agree |
| --- | --- | --- | --- | --- |
| □1 | □2 | □3 | □4 | □5 |

4. (Empathy) The answer provides emotional support and demonstrates medical responsibility.

| Strongly disagree | Disagree | Neither | Agree | Strongly agree |
| --- | --- | --- | --- | --- |
| □1 | □2 | □3 | □4 | □5 |

**eTable 5**. Statistical analysis of independent evaluations of 120 questions on the test set along four dimensions. A. Round 1. B. Round 2.

A

|  | Accuracy | Understandability | Trustworthiness | Empathy | Total Score | p value* |
|---|---|---|---|---|---|---|
| **Original** | 2.43 ± 1.17 | 2.25 ± 1.12 | 2.48 ± 1.38 | 2.15 ± 1.16 | 9.30 ± 4.42 | **NA** |
| **Role-play** | 3.17 ± 1.36 | 3.16 ± 1.45 | 3.34 ± 1.48 | 3.12 ± 1.46 | 12.79 ± 5.43 | **<0.001** |
| **Finetune 1** | 3.10 ± 1.27 | 3.26 ± 1.44 | 3.34 ± 1.54 | 3.25 ± 1.46 | 12.95 ± 5.41 | **<0.001** |
| **Finetune 2** | 3.09 ± 1.32 | 3.26 ± 1.43 | 3.33 ± 1.56 | 3.15 ± 1.48 | 12.83 ± 5.42 | **<0.001** |
| **Finetune 3** | <span style="color:red">3.38 ± 1.48</span> | <span style="color:red">3.46 ± 1.48</span> | <span style="color:red">3.50 ± 1.57</span> | <span style="color:red">3.45 ± 1.47</span> | <span style="color:red">13.79 ± 5.70</span> | **<0.001** |

B

|  | Accuracy | Understandability | Trustworthiness | Empathy | Total Score | p value* |
|---|---|---|---|---|---|---|
| **Best-finetune** | 3.00 ± 1.33 | 3.30 ± 1.41 | 3.26 ± 1.49 | 3.52 ± 1.49 | 13.08 ± 5.43 | NA |
| **Role-play+database** | 3.18 ± 1.21 | 3.56 ± 1.43 | 3.44 ± 1.45 | 3.51 ± 1.35 | 13.69 ± 5.15 | 0.351 |
| **Best-finetune+database** | 3.18 ± 1.14 | 3.56 ± 1.35 | 3.47 ± 1.36 | 3.51 ± 1.30 | 13.73 ± 4.83 | 0.458 |
| **Role-play+book** | 3.32 ± 1.17 | 3.67 ± 1.29 | 3.63 ± 1.32 | 3.79 ± 1.22 | 14.41 ± 4.75 | 0.084 |
| **Best-finetune+book** | <span style="color:red">3.44 ± 1.19</span> | <span style="color:red">3.88 ± 1.30</span> | <span style="color:red">3.81 ± 1.30</span> | <span style="color:red">4.01 ± 1.17</span> | <span style="color:red">15.14 ± 4.64</span> | **0.001** |

*Each round of each adjusted model and the base model (Llama2 or Best-finetune) was compared separately. P value is calculated with Mann-Whitney U test, with significant differences indicated in bold. The best score in each column is shown in red. Original (Llama2), Role-play (Original plus Role-play), Finetune 1-3 (Finetuned model versions 1-3 plus Role-play), Best-finetune (Finetune 3), Role-play+database (Role-play plus manual database retrieval), Best-finetune+database (the best finetuned model plus manual database retrieval), Role-play+book (Role-play plus book retrieval), Best-finetune+book (the best finetuned model plus book retrieval).

**eTable 6**. Inter-rater reliability analysis of 120 questions on the test set along four dimensions. A. Independent evaluation. B. Pair ranking evaluation.

**A**

|  | Round 1 | | | Round 2 | | |
|---|---|---|---|---|---|---|
|  | Rater 1 Mean ± SD | Rater 2 Mean ± SD | Kappa | Rater 1 Mean ± SD | Rater 2 Mean ± SD | Kappa |
| **Accuracy** | 3.09 ± 1.42 | 2.98 ± 1.37 | 0.711 | 3.03 ± 1.39 | 2.94 ± 1.33 | 0.861 |
| **Understandability** | 3.15 ± 1.52 | 3.00 ± 1.42 | 0.767 | 3.36 ± 1.56 | 3.26 ± 1.50 | 0.872 |
| **Trustworthiness** | 3.29 ± 1.62 | 3.10 ± 1.55 | 0.692 | 3.31 ± 1.59 | 3.19 ± 1.50 | 0.839 |
| **Empathy** | 3.16 ± 1.58 | 2.89 ± 1.45 | 0.726 | 3.60 ± 1.63 | 3.18 ± 1.46 | 0.611 |

**B**

|  | Rater 3 N (%) * | Rater 4 N (%) * | Kappa |
|---|---|---|---|
| **Accuracy** | 94 (78.3) | 86 (71.7) | 0.871 |
| **Understandability** | 73 (60.8) | 55 (45.8) | 0.732 |
| **Trustworthiness** | 47 (39.2) | 51 (42.5) | 0.855 |
| **Empathy** | 14 (11.7) | 12 (10.0) | 0.833 |

*Number and percentage of responses where ophthalmologists outperform EyeGPT were provided. SD=standard deviation.

**eTable 7.** Examples of generated answers with difference grating in independent evaluation.

| Question [Character] | Answer [Model] | Accuracy | Understandability | Trustworthiness | Empathy | Note | |
|---|---|---|---|---|---|---|---|
| What are the differences between rhegmatogenous, tractional, and exudative retinal detachments? [Medical students] | Great question! Retinal detachment is a serious condition that can cause vision loss if not treated promptly. Here are the differences between rhegmatogenous, tractional, and exudative retinal detachments: Rhegmatogenous retinal detachment occurs when there is a tear or hole in the retina, allowing fluid to pass through and separate the retina from the underlying tissue. This type of detachment is often caused by aging, injury, or diseases such as diabetes. Tractional retinal detachment occurs when scar tissue forms on the surface of the retina, pulling it away from the underlying tissue. This can be caused by conditions such as uveitis, tumors, or inflammation. Exudative retinal detachment occurs when fluid leaks out of the blood vessels in the retina, causing separation from the underlying tissue. This type of detachment is often caused by diseases such as age-related macular degeneration or diabetic retinopathy. In all cases, prompt medical attention is necessary to prevent permanent vision loss. As an ophthalmologist, I would perform a comprehensive eye exam to diagnose the type of retinal detachment and develop a treatment plan tailored to the individual patient's needs. Treatment may include surgery, laser therapy, or other interventions to repair the retina and prevent further vision loss. I hope this information helps you understand the differences between these types of retinal detachments. Let me know if you have any other questions! [Best-finetune+book] | 5 | 5 | 5 | 5 | ● | The answer accurately explained the concepts of rhegmatogenous, tractional, and exudative retinal detachment, including their causes and mechanisms. |
| | | | | | | ● | The answer was concise and appropriately uses medical terminology, making it suitable for medical students studying the subject. |
| | | | | | | ● | The answer was reliable and does not contain any unsafe or biased information. |
| | | | | | | ● | The answer could address any further questions, demonstrating a caring attitude towards medical students seeking information. |
| My child has been diagnosed with morning glory syndrome. Is | Morning glory syndrome is a rare condition, affecting approximately 1 in 100,000 individuals worldwide. It is characterized by a congenital abnormality of the retina, where the blood vessels in the retina are abnormally dilated, giving the appearance of a morning glory flower. While it is uncommon, it is important to seek medical attention if you notice any changes in your child's vision or if they exhibit any signs | 4 | 4 | 4 | 4 | ● | The answer mentioned the rarity of Morning Glory syndrome and the congenital abnormalities of the retina but incorrectly referred to it as vascular dilation instead of optic nerve dilation. |

| Question | Answer | | | | | Evaluation |
|---|---|---|---|---|---|---|
| morning glory syndrome common? [Patients] | of the condition, such as flashes of light or floaters. As an ophthalmologist, I understand the importance of early diagnosis and treatment to prevent long-term vision loss. [Role-play+book] | | | | | • The answer was clear, but it could be made more concise and easily understandable when explaining to the patient.<br>• The answer provided some safe and accurate medical information, but it omitted information about optic nerve damage and failed to provide warnings or preventive measures regarding patient safety and potential biases.<br>• The answer emphasized the importance of early diagnosis and treatment but lacked emotional resonance with parents and a deeper level of care and understanding. |
| Can hormones be used for CSC? [Medical students] | As an ophthalmologist, I understand that you are asking about the potential use of hormones in treating central serous chorioretinopathy (CSC). While there is some evidence to suggest that hormonal changes may play a role in the development of CSC, there is currently no established treatment that utilizes hormones to manage the condition. [Best-finetune+database] | 3 | 3 | 3 | 3 | • The answer addressed the uncertainty of using hormone therapy for CSC, but it did not specifically explain the necessity of discontinuing hormone use.<br>• The answer was concise and clear, but there is room for improvement in emphasizing key points.<br>• The answer failed to mention the safety risks associated with the use of hormones, which is an important consideration. Mishandling can potentially pose safety risks to clinical treatment.<br>• Additional background information or recommendations about CSC could be provided to meet the knowledge needs of medical students. |
| Is Leber hereditary optic neuropathy inherited? [Patients] | Leber hereditary optic neuropathy (LHON) is a rare genetic disorder that affects the optic nerve and can cause vision loss. It is caused by mutations in the genes that code for the proteins involved in the metabolism of the amino acid histidine. These mutations are inherited in an autosomal dominant pattern, which means that a single copy of the mutated gene is enough to cause the condition. This means that if one of your parents carries the mutated gene, you have a 50% chance of inheriting the mutated gene and developing LHON. However, it's important to note that not everyone who inherits the mutated gene will develop LHON, and some people may not develop any symptoms until later in life. Additionally, LHON can | 2 | 2 | 2 | 2 | • The answer only indicated the hereditary nature of LHON but did not accurately describe the genetic pattern of LHON.<br>• The answer's structure was relatively clear, but it was not concise. Complex terminology may make it difficult for ordinary readers to understand.<br>• The answer included inaccurate key medical information, which may mislead users seeking accurate information. |

| | | | | | |
|---|---|---|---|---|---|
| | also be caused by acquired mutations, which can occur due to various factors such as viral infections or exposure to certain toxins.<br>[Finetune 1] | | | | • The answer did not emphasize the importance of timely medical and genetic counseling, lacking care and understanding for the patients. |
| What characterizes PACG?<br>[Medical students] | Here are some key characteristics of PACG:<br>1. **Deck-building:** Players start with a small deck of cards and must strategically add new cards to their deck throughout the game to improve their chances of winning.<br>2. **Hand management:** Each player has a limited number of hands they can play each turn, and they must carefully choose which cards to play and when.<br>3. **Card combinations:** Cards can be combined in various ways to create powerful combinations and strategies.<br>4. **Interaction:** Players can interact with each other in various ways, such as by playing cards on each other's turns or by using special abilities to disrupt their opponents' plans.<br>5. **Luck vs. strategy:** While luck can play a role in the game; it is ultimately a game of strategy and skill. Players must carefully plan their deck-building and card-playing strategies to come out on top.<br>6. **Replay ability:** With a large deck of cards and a variety of playstyles, each game of PACG can be different and offer new challenges and opportunities for strategic play.<br>7. **Scalability:** PACG can be played with 2-4 players, making it a great option for groups of any size.<br>8. **Easy to learn:** While the game has a lot of depth and complexity, it is relatively easy to learn and can be taught to new players in a short amount of time.<br>9. **Accessible:** PACG is a game that can be enjoyed by players of all skill levels and ages, making it a great option for families and casual gamers.<br>10. **Expansions:** There are many expansions available for PACG, which offer new cards, mechanics, and gameplay elements to keep the game fresh and exciting. By understanding these characteristics, you should have a better idea of what to expect when playing PACG and how to approach the game."<br>[Original] | 1 | 1 | 1 | 1 | This answer did not correctly recognize the true meaning of PACG, resulting in the entire answer being completely unrelated to medicine. |

**eTable 8.** Pairwise ranking study of EyeGPT(best) answers and ophthalmologist answers.

|  | EyeGPT (best) | Ophthalmologist | Tie | p value* |
|---|---|---|---|---|
| **Accuracy** | 3 (2.5%) | 85 (70.8%) | 32 (26.7%) | **0.037** |
| **Understandability** | 2 (1.7%) | 54 (45.0%) | 64 (53.3%) | 0.566 |
| **Trustworthiness** | 2 (1.7%) | 44 (36.7%) | 74 (61.7%) | 0.730 |
| **Empathy** | 1 (0.8%) | 11 (9.2%) | 108 (90.0%) | 1.000 |

*P value is calculated with chi-square test, with significant differences indicated in bold.

EyeGPT(best)=the Best-finetune+book model.

**eTable 9.** Examples in pairwise ranking evaluation. A. The specific answers of EyeGPT(best) comparison to Ophthalmologist. B. The full set of ratings for each answer in A.

A

| Question | EyeGPT(best) answer | Ophthalmologist answer |
|---|---|---|
| Q1: Is myopia reversible? [Patients] | Great question! Myopia, or nearsightedness, can be a complex condition, and the answer to this question depends on several factors. Firstly, it is important to understand that myopia is not a disease or an injury, but rather a refractive error caused by the shape of the cornea or lens of the eye. This means that the elasticity of the cornea or lens is not perfect, causing light rays to focus in front of the retina instead of directly on it. Now, regarding reversibility, the good news is that myopia can be corrected with glasses, contact lenses, or refractive surgery. These treatments work by changing the way light rays enter the eye, allowing them to focus directly on the retina. However, the degree of correction may vary depending on the severity of the myopia. However, it is important to note that myopia is a progressive condition, which means that it can worsen over time if left untreated. This means that even if you correct your myopia with glasses or contact lenses, the degree of correction may increase over time as your myopia progresses. In summary, myopia is reversible to some extent, but it is a progressive condition that can worsen over time if left untreated. Regular eye exams are important to monitor and manage your myopia to ensure the best possible vision and to prevent any potential complications. Do you have any other questions about myopia or eye care in general? | I'm sorry to inform you that myopia is not reversible. Axial myopia is caused by the elongation of the eyeball, and in general, the length of the eyeball is difficult to change, making myopia difficult to reverse. However, the progression of myopia can sometimes naturally stabilize or slow down as a person reaches adulthood. Additionally, certain treatments and interventions that I mentioned earlier can help manage myopia and potentially slow its progression. |
| Q2: I have been diagnosed with retinal detachment. What is retinal | Retinal detachment is a medical condition where the retina, the layer of tissue at the back of the eye that senses light and sends visual signals to the brain, becomes separated from the underlying tissue. This can cause vision loss or blindness if left untreated. Retinal detachment occurs when the retina becomes damaged or weakened, allowing the retinal | Hello, I understand that being diagnosed with a condition like retinal detachment can be quite overwhelming. Let me explain this disease to you. The retina is a thin layer of tissue that lines the back of the eye on |
| detachment? [Patients] | cells to pull away from the underlying tissue. This can be caused by a variety of factors, including age-related changes, injury, disease, or certain medical conditions. Symptoms of retinal detachment can include sudden flashes of light, blurred vision, distorted vision, or loss of vision in one or more areas of vision. If you experience any of these symptoms, it is important to seek medical attention immediately, as prompt treatment can help to preserve vision and prevent permanent damage. In diagnosing retinal detachment, an eye doctor will perform a comprehensive eye exam to assess the retina and detect any signs of detachment. Treatment options for retinal detachment can include laser surgery, photocoagulation, or vitrectomy, depending on the severity of the detachment and the underlying cause. It is important to note that retinal detachment is a medical emergency, and prompt treatment is essential to preserve vision and prevent permanent damage. If you suspect you or someone you know may have retinal detachment, do not hesitate to seek medical attention from an eye doctor or a medical professional. | the inside. It is responsible for receiving light that the lens has focused, converting the light into neural signals, and sending these signals on to the brain. In simpler terms, you can think of the retina as the "camera film" of the eye. Retinal detachment is a serious condition where the retina pulls away from its normal position at the back of the eye. This separation prevents the retina from correctly interacting with the choroid, a layer of blood vessels that supplies the retina with oxygen and nutrients, and can cause permanent vision loss if not treated promptly. Symptoms of retinal detachment often include: - Sudden appearance of floaters – small specks or threads that seem to float through your field of vision. - Flashes of light in one or both eyes. - Gradual or sudden decrease in vision, particularly if it's like a curtain being drawn across your field of vision. Treatment usually involves surgery. The prognosis can be good if treated early, but vision may not return to normal if the macula (central part of the retina responsible for detailed vision) has been detached. |
| Q3: What are the potential complications of Stickler syndrome? [Medical students] | Stickler syndrome is a rare genetic disorder that affects the connective tissue, including the eyes. The potential complications of Stickler syndrome in the eyes can include: 1. Retinal detachment: This is a common complication of Stickler syndrome, and it occurs when the retina becomes separated from the underlying tissue. Retinal detachment can cause vision loss and blindness if left untreated. | Stickler syndrome is a multisystem condition, and its complications can affect various parts of the body, including the eyes, ears, joints, and skeletal system. I will only describe the complications associated with ophthalmology here. ● Retinal Detachment: This is a significant risk for individuals |

2. Macular hole: A macular hole is a small defect in the central part of the retina, which can cause vision loss. Macular holes can be caused by Stickler syndrome and can lead to blindness if left untreated.

3. Vitreomacular traction: This is a condition where the vitreous gel pulls on the retina, causing vision loss. Vitreomacular traction can be caused by Stickler syndrome and can lead to blindness if left untreated.

4. Glaucoma: Stickler syndrome can increase the risk of developing glaucoma, a group of eye conditions that can damage the optic nerve and cause vision loss.

5. Cataracts: Stickler syndrome can increase the risk of developing cataracts, which are cloudy areas in the lens of the eye that can cause vision loss.

6. Optic nerve damage: Stickler syndrome can cause damage to the optic nerve, which can lead to vision loss and blindness.

7. Eye movement problems: Stickler syndrome can cause problems with eye movements, such as nystagmus (involuntary eye movements) or ophthalmoplegia (paralysis of the eye muscles).

8. Proptosis (bulging eyes): Stickler syndrome can cause the eyes to protrude abnormally, which can be a sign of the disorder.

9. Eye infections: Stickler syndrome can increase the risk of developing eye infections, such as conjunctivitis (pink eye) or endophthalmitis (inflammation of the eye).

10. Blindness: In severe cases, Stickler syndrome can lead to blindness.

It's important to note that not everyone with Stickler syndrome will experience these complications with Stickler syndrome and can lead to vision loss if not promptly treated.

- Myopia: High degrees of nearsightedness are common and can increase the risk of retinal issues.
- Cataracts: Early-onset cataracts can develop, sometimes leading to impaired vision.
- Glaucoma: Increased intraocular pressure can occur, potentially damaging the optic nerve.
- Vitreous Anomalies: Abnormalities in the vitreous gel of the eye can lead to visual problems.

B

| Question [Character] | Q1: Is myopia reversible? [Patients] | Q2: I have been diagnosed with retinal detachment. What is retinal detachment? [Patients] | Q3: What are the potential complications of Stickler syndrome? [Medical students] |
|---|---|---|---|
| **Accuracy** | Ophthalmologist | Tie | EyeGPT(best) |
| **Understandability** | Ophthalmologist | Ophthalmologist | EyeGPT(best) |
| **Trustworthiness** | Tie | Tie | EyeGPT(best) |
| **Empathy** | EyeGPT(best) | Ophthalmologist | Tie |